\documentclass{article}
\usepackage{geometry}
\usepackage{authblk}

\usepackage{hyperref}       
\usepackage{url}            
\usepackage{booktabs}       
\usepackage{amsfonts}       
\usepackage{nicefrac}       
\usepackage{microtype}      
\usepackage{natbib}

\usepackage{amsmath}
\usepackage{amsthm}
\usepackage{nccmath}
\usepackage{amsfonts}
\usepackage{amssymb}
\usepackage{bm}
\usepackage{bbm}
\usepackage{mathtools}

\usepackage[dvipsnames]{xcolor}
\usepackage{pgfplots}
\usepackage{tikz}
\tikzstyle{every picture}+=[remember picture]
\usetikzlibrary{arrows,automata,calc}
\pgfplotsset{compat=newest}

\usepackage{graphicx}
\usepackage{subcaption}
\usepackage{enumitem}
\usepackage{xfrac}
\usepackage{algorithm}                                                          
\usepackage{algorithmic}

\usepackage{xspace}                                                             
                                      
\DeclareRobustCommand{\ie}{i.e.,\@\xspace}                                      
  
\renewcommand{\Re}{\mathbb R}

\newcommand{\transp}{\mathsf{T}}

\newcommand{\wh}[1]{\widehat{#1}}

\usepackage{accents}
\DeclareMathAccent{\wtilde}{\mathord}{largesymbols}{"65}

\DeclareRobustCommand{\ie}{i.e.,\@\xspace}

\DeclareRobustCommand{\st}{s.t.\@\xspace}

\newlength{\minipagewidth}
\newlength{\minipagewidthx}
\usepackage[colorinlistoftodos, textwidth=20mm]{todonotes}
\definecolor{citrine}{rgb}{0.89, 0.82, 0.04}
\definecolor{blued}{RGB}{70,197,221}

\newtheorem{theorem}{Theorem}
\newtheorem{lemma}[theorem]{Lemma}
\newtheorem{proposition}[theorem]{Proposition}

\theoremstyle{remark}

\theoremstyle{theorem}

\title{Concentration Inequalities for Multinoulli Random Variables}

\author[1]{Jian Qian}
\author[1]{Ronan Fruit}
\author[1]{Matteo Pirotta}
\author[2]{Alessandro Lazaric}
\affil[1]{Sequel Team - Inria Lille}
\affil[2]{Facebook AI Research}

\date{July 2018}

\begin{document}
\maketitle

\begin{abstract}
        We investigate concentration inequalities for Dirichlet and Multinomial random variables.
\end{abstract}

\section{Problem Formulation}
We analyse the concentration properties of the random variable $Z_n \geq 0$ defined as:
\begin{equation}\label{eq:Zn_noncentered}
        Z_n := \max_{v \in [0,D]^S} \left\{ \left(\wh{p}_n - p \right)^\transp v \right\}
\end{equation}
where $\wh{p}_n \in \Delta^{S}$ is a random vector, $p \in \Delta^S$ is deterministic and $\Delta^S = \{x \in \Re^S ~:~ \sum_{i=1}^S x_i = 1 \wedge x_i \geq 0 \}$ is the $(S-1)$-dimensional simplex.
It is easy to show that the maximum in Eq.~\ref{eq:Zn_noncentered} is equivalent to computing the (scaled) $\ell_1$-norm of the vector $\wh{p}_n - p$:
\begin{equation}\label{eq:Zn_l1norm}
        Z_n = \max_{u \in [-\frac{D}{2}, \frac{D}{2}]} \left\{ (\wh{p}_n - p)^\transp \left( u + \frac{D}{2}e \right) \right\}
            = \frac{D}{2} \| \wh{p}_n - p \|_1
\end{equation}
where we have used the fact that $\frac{D}{2}(\wh{p}_{n} - p)^\transp e = 0$.
As a consequence, $Z_n$ is a bounded random variable in $[0,D]$.
While the following discussion apply to Dirichlet distributions, we focus on $\wh{p}_n \sim \frac{1}{n}\textit{Multinomial}(n,p)$. The results previously available in the literature are summarized in the following.

The literature has analysed the concentration of the $\ell_1$-discrepancy of the true distribution and the empirical one in this setting.
\begin{proposition}{\citep{weissman2003inequalities}}\label{lem:multinomial.weissman}
        Let $p \in \Delta^S$ and $\wh{p} \sim \frac{1}{n} \textit{Multinomial}(n, p)$. Then, for any $S \geq 2$ and $\delta \in [0,1]$:
        \begin{equation}\label{eq:multinomial.weissman}
               \mathbb{P}\Bigg( \|\wh{p} - p \|_1 \geq \sqrt{\frac{2S\ln(\sfrac{2}{\delta})}{n}} \Bigg) \leq
               \mathbb{P}\Bigg( \|\wh{p} - p \|_1 \geq \sqrt{\frac{2 \ln\left(\sfrac{(2^S-2)}{\delta}\right)}{n}} \Bigg) \leq \delta 
        \end{equation}
\end{proposition}
This concentration inequality is at the core of the proof of UCRL, see~\citep[][App. C.1]{Jaksch10}.
Another inequality is provided in~\citep[][Lem. 3]{devroye1983l1}.
\begin{proposition}{\citep{devroye1983l1}}\label{lem:multinomial.devroye}
        Let $p \in \Delta^S$ and $\wh{p} \sim \frac{1}{n} \textit{Multinomial}(n, p)$. Then, for any $0\leq \delta \leq 3 \exp \left(-4S/5 \right)$:
\begin{equation}\label{eq:multinomial.devroye}
        \mathbb{P}\Bigg( \|\wh{p}_n - p \|_1 \geq 5\sqrt{\frac{\ln(\sfrac{3}{\delta})}{n}} \Bigg) \leq \delta
\end{equation}
\end{proposition}
While Prop.~\ref{lem:multinomial.weissman} shows an explicit dependence on the dimension of the random variable, such dependence is hidden in Prop.~\ref{lem:multinomial.devroye} by the constraint on $\delta$.
Note that for any $0 \leq \delta \leq 3 \exp \left(-4S/5 \right)$, $\sqrt{\frac{\ln(\sfrac{3}{\delta})}{n}} > \sqrt{\frac{4S}{5n}}$.
This shows that the $\ell_1$-deviation always scales proportionally to the dimension of the random variable, \ie as $\sqrt{S}$.

\textit{A better inequality.}
The natural question is whether is possible to derive a concentration inequality independent from the dimension of $p$ by exploiting the correlation between $\wh{p}$ and the maximizer vector $v^*$.
This question has been recently addressed in~\citep[][Lem. C.2]{Agrawal2017posterior}:
\begin{lemma}{\citep{Agrawal2017posterior}}\label{lem:multinomial.agrawal}
Let $p \in \Delta^S$ and $\wh{p} \sim \frac{1}{n} \textit{Multinomial}(n, p)$. Then, for any $\delta \in [0,1]$:
\begin{equation*}
    \mathbb{P}\Bigg( \|\wh{p}_n - p \|_1 \geq \sqrt{\frac{2\ln(\sfrac{1}{\delta})}{n}} \Bigg) \leq \delta
\end{equation*}
\end{lemma}
Their results resemble the one in Prop.~\ref{lem:multinomial.devroye} but removes the constraint on $\delta$.
As a consequence, the implicit or explicit dependence on the dimension $S$ is removed.
In the following, we will show that Lem.~\ref{lem:multinomial.agrawal} may not be correct.

\section{Theoretical Analysis (the asymptotic case)} \label{sec:asymptotic}
In this section, we provide a counter-argument to the Lem.~\ref{lem:multinomial.agrawal} in the asymptotic regime (\ie $n \to +\infty$). The overall idea is to show that  the expected value of $Z_n$ asymptotically grows as $O(\sqrt{S})$ and $Z_n$ itself is well concentrated around its expectation. As a result, we can deduce that all quantiles of $Z_n$ grow as $O(\sqrt{S})$ as well.

We consider the true vector $p$ to be uniform, \ie $p = (\frac{1}{S}, \ldots, \frac{1}{S})$ and $\wh{p} \sim \frac{1}{n} \textit{Multinomial}(n,p)$.\footnote{The analysis holds also in the case $\wh{p} \sim \textit{Direchlet}(np)$, see~\citep{Osband2017posterior}.}
The following lemma provides a characterization of the variable $Z_S := \lim_{n\to +\infty} \sqrt{n}Z_n$.

\begin{lemma} \label{lem:Z_S.distribution}
        Consider $S\in \mathbb{N}$, $\mathcal{S} = \{1,\ldots,S\}$ and $p = (\frac{1}{S}, \ldots,\frac{1}{S})$ be the uniform distribution on $\mathcal{S}$.
        Let $e_S$ be the vector of ones of dimension $S$. Define $Y\sim \mathcal{N}(0,I_S-\frac{1}{S-1}N)$ where $N = e_S e_S^\transp - I_S$ is the matrix with $0$ in all the diagonal entry and $1$ elsewhere,
        and  $Y^+ = (\max (Y_i , 0))_{i\in \mathcal{S}}$. Then:
        \[
                Z_S = \lim_{n \to +\infty} \sqrt{n} Z_n \sim \| Y^+ \|_1 D \sqrt{\frac{S-1}{S^2}}.
        \]
Furthermore,
\begin{align*}
        \mathbb{E}[Z_S] = \sqrt{\frac{S-1}{S^2}}\cdot \mathbb{E}\left[  \sum_{i=1}^S Y_i^+ \right] = \sqrt{S-1} \cdot \mathbb{E}[Y_1^+] =  \sqrt{\frac{S-1}{2\pi}}.
\end{align*}
\end{lemma}

While the previous lemma may already suggest that $Z_S$ should grow as $O(\sqrt{S})$ as its expectation, it is still possible that a large part of the distribution is concentrated around a value independent from $S$, with limited probability assigned to, e.g., values growing as $O(S)$, which could justify the $O(\sqrt{S})$ growth of the expectation. Thus, in order to conclude the analysis, we need to show that $Z_S$ is concentrated ``enough'' around its expectation.

Since the random variables $Y_i$ are correlated, it is complicated to directly analyze the deviation of $Z_S$ from its mean. Thus we first apply an orthogonal transformation on $Y$ to obtain independent r.v. (recall that jointly normally distributed variables are independent if uncorrelated).

\begin{lemma} \label{lem:Y.normalization}
        Consider the same settings of Lem.~\ref{lem:Z_S.distribution} and recall that $Y\sim \mathcal{N}(0,I_S-\frac{1}{S-1}N)$.
        There exists an orthogonal transformation $U \in O_S(\mathbb{R})$, \st
                        $$W=\sqrt{\frac{S-1}{S}}UY \sim \mathcal{N}\left(0, \begin{bmatrix}
                                        I_{S-1} & 0 \\
                                        0 & 0    
                                \end{bmatrix}
                        \right).$$
\end{lemma}

By exploiting the transformation $U$ we can write that $Z_S \sim g(W) := \frac{1}{\sqrt{S}}e_S^\transp \left(U^\transp W\right)^+$.
Since $W_i$ are i.i.d. standard Gaussian random variables and $g$ is $1$-Lipschitz, we can finally characterize the mean and the deviations of $Z_S$ and derive the following anticoncentration inequality for $Z_S$.

\begin{theorem} \label{thm:anticoncentration}
Let $p \in \Delta^{S} = (\frac{1}{S}, \ldots, \frac{1}{S})$ and $\wh{p}_n \sim \frac{1}{n} \textit{Multinomial}(n, p)$.
Define $Z_n = \max_{v \in [0,D]} \left\{ (\wh{p}_n - p)^\transp v \right\}$ and $Z_S = \lim_{n \to +\infty} \sqrt{n} Z_n$.
Then, for any $\delta \in (0,1)$:
\[
        \mathbb{P} \Big[ Z_S \geq \sqrt{\frac{2(S-1)}{\pi}} - \sqrt{2\log(2/\delta)} \Big] \geq 1-\delta.
\]
\end{theorem}

This result shows that every quantile of $Z_S$ is dependent on the dimension of the random variable, \ie $\sqrt{S}$.
Similarly to Lem.~\ref{lem:multinomial.devroye}, it is possible to lower bound the quantile by a dimension-free quantity at the price of having an exponential dependence on $S$ in $\delta$.

\clearpage
\newpage

\appendix
\section{Proof for the asymptotic scenario}
\label{app:asymptotic}
In this section we report the proofs of lemmas and theorem stated in Sec.~\ref{sec:asymptotic}.

\subsection{Proof of Lem.~\ref{lem:Z_S.distribution}}
Let $Y_{n,i} = \frac{1}{\sqrt{n}\sqrt{\frac{S-1}{S^2}}}\sum\limits_{j=1}^{n}(X_i^j- \frac{1}{S})$ and $Y_n = (Y_{n,i})_{i\in \mathcal{S}}$. Then:
\begin{align*}
\sqrt{n}Z_n 
&= \sqrt{n} \max_{v\in[0,D]^S}(\wh{p}-p)^\transp v 
=\sqrt{n} \max_{v\in[0,D]^S} \sum\limits_{i=1}^{S} \frac{v_i}{n} \sum\limits_{j=1}^{n}(X_i^j - \frac{1}{S})\\
&= \max\limits_{v\in[0,D]^S} \sum\limits_{i=1}^{S} Y_{n,i}v_i \sqrt{\frac{S-1}{S^2}}
 = D \sqrt{\frac{S-1}{S^2}} \cdot e^\transp Y_{n}^+,
\end{align*}
where we used the fact that the $v$ maximizing $Z_n$ takes the largest value $D$ for all positive components $Y_{n,i}$ and is equal to $0$ otherwise.
We recall that the covariance of the normalized multinoulli variable $Y_{n,i}$ with probabilities $p_i = 1/S$ is $I_S-\frac{1}{S-1}N$. As a result, a direct application of the central limit theorem gives $Y_n \stackrel{\mathcal{D}}{\to} Y\sim \mathcal{N}(0,I_S-\frac{1}{S-1}N)$. Then we can apply the functional CLT and obtain $Z_S = \lim\limits_{n\to \infty}\sqrt{n}Z_n = \lim\limits_{n\to \infty} e_S^\transp  Y_n^+  \sqrt{\frac{S-1}{S^2}}
 \stackrel{\mathcal{D}}{\sim} \sqrt{\frac{S-1}{S^2}} \cdot e_S^\transp Y^+$, where $Y^+$ is a random vector obtained by truncating from below at $0$ the multi-variate Gaussian vector $Y$. Since the marginal distribution of each random variable $Y_i$ is $\mathcal{N}(0,1)$, \ie are identically distributed (see definition in Lem.~\ref{lem:Z_S.distribution}), $Y^+_i$ has a distribution composed by a Dirac distribution in $0$ and a half normal distribution, and its expected value is $\mathbb{E}[Y^+_i] = 1/\sqrt{2\pi}$, while leads to the final statement on the expectation.

\subsection{Proof of Lem.~\ref{lem:Y.normalization}}
        Denote $\lambda(A)$ the set of eigenvalues of square matrix $A$.
        Let $B \in \Re^{S \times S}$ such that $B = \begin{bmatrix}
0_{S,S-1} & e_{S}
\end{bmatrix}$, where $0_{S,S-1} \in \Re^{S\times (S-1)}$ is a matrix full of zeros. Then, we can write the eigenvalues of the covariance matrix of $Y$ as
\begin{align*}
        \lambda(I_S-\frac{1}{S-1}N) & = \lambda(\frac{S}{S-1} I_S - \frac{1}{S-1}e_S e_S^\transp)
                                      = \lambda \left( \frac{S}{S-1}I_S - \frac{1}{S-1} B B^\transp\right)\\
                                    & =\frac{S}{S-1} \lambda\left(I_S - \frac{1}{S-1} B^\transp B \right) 
                                      = \frac{S}{S-1} \lambda\left(I_S - 
                                      \begin{bmatrix}
                                              0_{S-1} & 0 \\
                                              0 & 1
                                      \end{bmatrix}\right),
\end{align*}
where we use the fact that $\lambda(I-A^\transp A) = \lambda(I-AA^\transp)$. As a result, the covariance of $Y$ has one eigenvalue at $0$ and eigenvalues equal to $\frac{S}{S-1}$ with multiplicity $S-1$. As a result, we can diagonalize it with an orthogonal matrix $U\in O_S(\mathbb{R})$ (obtained using the normalized eigenvectors) and obtain
\begin{align*}
U (I_S-\frac{1}{S-1}N) U^T = \begin{bmatrix}
\frac{S}{S-1}I_{S-1} & 0 \\
0 & 0    
\end{bmatrix}.
\end{align*}
Define $W = \sqrt{\frac{S-1}{S}} UY$, then:
\begin{align*}
Cov(W,W) &= \frac{S-1}{S}Cov(UY,UY) = \frac{S-1}{S}U Cov(Y,Y)U^T\\
         &= \frac{S-1}{S}U (I_S-\frac{1}{S-1}N) U^T 
          = \begin{bmatrix}
I_{S-1} & 0 \\
0 & 0    
\end{bmatrix}.
\end{align*}
Thus $W\sim  \mathcal{N}\left(0, \begin{bmatrix}
I_{S-1} & 0 \\
0 & 0    
\end{bmatrix}
\right)$.

\subsection{Proof of Thm.~\ref{thm:anticoncentration}}

By exploiting Lem.~\ref{lem:Z_S.distribution} and Lem.~\ref{lem:Y.normalization} we can write:
\begin{align*}
Z_S  & \sim e_S^\transp Y^+ \cdot  \sqrt{\frac{S-1}{S^2}} 
     = e_S^\transp \left(\sqrt{\frac{S}{S-1}} U^\transp W \right)^+ \cdot \sqrt{\frac{S-1}{S^2}}
     = e_S^\transp \left( U^\transp W \right)^+ \cdot \frac{1}{\sqrt{S}}
\end{align*}
Let $g(\cdot) = e_S^\transp \left(U^T\cdot \right)^+  \frac{1}{\sqrt{S}}$. Then $g$ is $1$-Lipschitz:
\begin{align*}
        |g(x) - g(y)| &\leq Lip(e_s^\transp \cdot) Lip(U^\transp\cdot) Lip((\cdot)^+) \frac{1}{\sqrt{S}} \|x - y\|_2
        = \sqrt{S} \cdot 1 \cdot 1 \cdot \frac{1}{\sqrt{S}} \| x- y\|_2
\end{align*}
where $Lip(f)$ denotes the Lipschitz constant of a function $f$ and we exploit the fact that $U$ is an orthonormal matrix.

We can now study the concentration of the variable $Z_S$.
Given that $W$ is a vector of i.i.d. standard Gaussian variables\footnote{Note that we can drop the last component of $W$ since it is deterministically zero.} and $g$ is $1$-Lipschitz, we can use~\citep[][Thm. 2.4]{wainwright2017concetration} to prove that for all $t > 0$:
\begin{align*}
\mathbb{P}(Z_S \geq \mathbb{E}[Z_S] - t) &\geq 1-\mathbb{P}(|Z_S- \mathbb{E}[Z_S]|\geq t) \geq 1- 2 e^{-\frac{t^2}{2}}.
\end{align*}
Substituting the value of $\mathbb{E}[Z_S]$ and inverting the bound gives the desired statement.


\bibliographystyle{plainnat}
\bibliography{span}

\begin{thebibliography}{6}
\providecommand{\natexlab}[1]{#1}
\providecommand{\url}[1]{\texttt{#1}}
\expandafter\ifx\csname urlstyle\endcsname\relax
  \providecommand{\doi}[1]{doi: #1}\else
  \providecommand{\doi}{doi: \begingroup \urlstyle{rm}\Url}\fi

\bibitem[Agrawal and Jia(2017)]{Agrawal2017posterior}
Shipra Agrawal and Randy Jia.
\newblock Optimistic posterior sampling for reinforcement learning: worst-case
  regret bounds.
\newblock In \emph{{NIPS}}, pages 1184--1194, 2017.

\bibitem[Devroye(1983)]{devroye1983l1}
Luc Devroye.
\newblock The equivalence of weak, strong and complete convergence in $\ell_1$
  for kernel density estimates.
\newblock \emph{The Annals of Statistics}, 11\penalty0 (3):\penalty0 896--904,
  09 1983.
\newblock \doi{10.1214/aos/1176346255}.

\bibitem[Jaksch et~al.(2010)Jaksch, Ortner, and Auer]{Jaksch10}
Thomas Jaksch, Ronald Ortner, and Peter Auer.
\newblock Near-optimal regret bounds for reinforcement learning.
\newblock \emph{Journal of Machine Learning Research}, 11:\penalty0 1563--1600,
  2010.

\bibitem[Osband and Roy(2017)]{Osband2017posterior}
Ian Osband and Benjamin~Van Roy.
\newblock Why is posterior sampling better than optimism for reinforcement
  learning?
\newblock In \emph{{ICML}}, volume~70 of \emph{Proceedings of Machine Learning
  Research}, pages 2701--2710. {PMLR}, 2017.

\bibitem[Wainwright(2017)]{wainwright2017concetration}
Wainwright.
\newblock High-dimensional statistics: A non-asymptotic viewpoint.
\newblock 2017.

\bibitem[Weissman et~al.(2003)Weissman, Ordentlich, Seroussi, Verdu, and
  Weinberger]{weissman2003inequalities}
Tsachy Weissman, Erik Ordentlich, Gadiel Seroussi, Sergio Verdu, and Marcelo~J
  Weinberger.
\newblock Inequalities for the l1 deviation of the empirical distribution.
\newblock Technical Report HPL-2003-97R1, Hewlett-Packard Labs, 2003.

\end{thebibliography}

\end{document}